\documentclass[letterpaper, 10 pt, conference]{ieeeconf}

\IEEEoverridecommandlockouts                              

\usepackage{amsmath} 

\usepackage{graphicx}
\usepackage{caption}
\usepackage{subcaption}
\usepackage{hyperref}

\title{\LARGE \bf
Learning Precise 3D Manipulation from Multiple Uncalibrated Cameras
}

\author{Iretiayo Akinola$^{*,1}$, Jacob Varley$^{2}$, and Dmitry Kalashnikov$^{2}$
\thanks{$^{*}$Work done during internship at Robotics at Google, NYC}
\thanks{$^{1}$Columbia University}
\thanks{$^{2}$Robotics at Google}
}

\begin{document}

\maketitle
\thispagestyle{empty}
\pagestyle{empty}

\begin{abstract}
In this work, we present an effective multi-view approach to closed-loop end-to-end learning of precise manipulation tasks that are 3D in nature.
Our method learns to accomplish these tasks using multiple statically placed but uncalibrated RGB camera views without building an explicit 3D representation such as a pointcloud or voxel grid. This multi-camera approach achieves superior task performance on difficult stacking and insertion tasks compared to single-view baselines. Single view robotic agents struggle from occlusion and challenges in estimating relative poses between points of interest.
While full 3D scene representations (voxels or pointclouds) are obtainable from registered output of multiple depth sensors, several challenges complicate operating off such explicit 3D representations. These challenges include imperfect camera calibration, poor depth maps due to object properties such as reflective surfaces, and slower inference speeds over 3D representations compared to 2D images.
Our use of static but uncalibrated cameras does not require camera-robot or camera-camera calibration making the proposed approach easy to setup and our use of \textit{sensor dropout} during training makes it resilient to the loss of camera-views after deployment.
\end{abstract}


\section{INTRODUCTION}
Precise object manipulation remains an active area of robotics research;  it finds applications in diverse domains such as manufacturing robotics, warehouse packaging and home-assistant robotics. Until recently, most automated solutions are designed for and deployed to highly instrumented settings where scripted robot actions are repeated to move through predefined set of positions. This approach often requires a highly calibrated setup which can be expensive and time-consuming. Also, they lack robustness needed to handle changes in environment, and configuring such methods for new settings requires significant engineering efforts.
Advancements in computer vision have led to superior performances in robotic grasping in dense clutter \cite{kalashnikov2018qt}\cite{morrison2018closing}\cite{wu2019pixeliros}\cite{zeng2018robotic}\cite{zeng2018learning} by allowing robotic systems to make use of vision systems for various manipulation tasks in less structured settings.

While there has been recent progress in vision-based robot manipulation such as grasping, other tasks like stacking, insertion and precision kitting, that require precise object manipulation, remain challenging for robotic systems\cite{kroemer2019review}. These sorts of tasks require accurate 3D geometric knowledge of the task environment including object shape and pose, relative distances and orientation between key locations in the scene among others.
For example, solving an insertion task requires picking up an object using the geometry and pose of the object and sticking it in a hole using the pose of object relative to the hole.

\begin{figure}[t]
    \centering
    \includegraphics[width=0.95\linewidth]{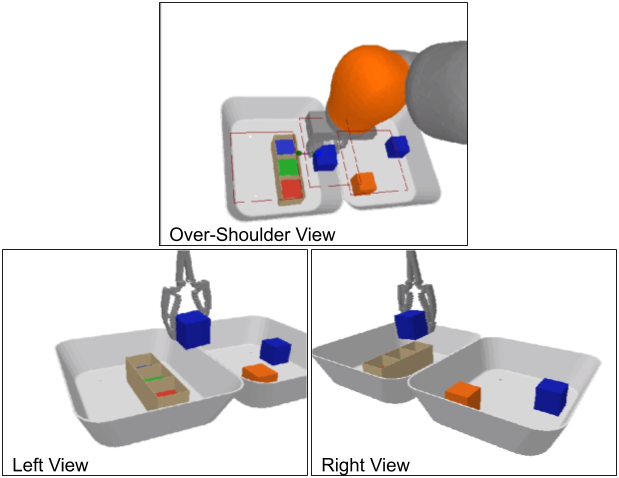}
    \caption{\small \textbf{Multi-view Task Learning.} An insertion task where a block is placed into a fixture. This task requires 3D understanding and alignment. A single view system that sees only one of the images would have a difficult time resolving the alignment challenge.  Our system combines information from multiple views and achieves better performance on precision-based robotic tasks.}
    \label{fig:three_view_alignment}
\vspace{-4mm}
\end{figure}

We observe that the majority of the existing reinforcement-learned vision-based robotic manipulation systems employ a single camera to observe the task scene. However, the rich 3D information required for solving precision-based tasks are usually limited from a single camera input. For example, it is usually hard to resolve scale and alignment from a single view. Even for humans, navigating a room or completing a task with one eye closed becomes more challenging from a lack of depth perception. In addition, single view systems are very susceptible to occlusion during task learning requiring the robot to actively move out of the way and reset during task execution. To address these limitations, we propose using multi-view camera setup such as that shown in Figure \ref{fig:three_view_alignment} to solve precision-based object manipulation. Since cameras are cheap and ubiquitous, adding a few more cameras to capture multiple views of the task scene is a practical and feasible option.

This research develops techniques for combining multiple camera views to improve the state estimation and increase the robustness of robot action in learning-based robotic manipulation systems.
Our approach is a reinforcement learning based method that takes in multiple color (RGB) images from different viewpoints as input and produces robot actions in a closed-loop fashion. The system is trained end-to-end.


Our key contributions include:
\begin{itemize}
    \item A novel camera calibration free, multi-view, approach to precise 3D robotic manipulation
    \item An RL model architecture featuring a \textit{sensor dropout} training regime that achieves large reductions to error rates on precision-based tasks (Stacking I \textbf{49.18\%}, Stacking II \textbf{56.84\%}, Insertion \textbf{64.1\%})
    \item Analysis of a number of model-free RL architectures for efficiently learning precision based robotics from individual, depth, and multiple views
\end{itemize}

Video of this paper can be found at \url{https://www.youtube.com/watch?v=02dhUfTJNK4}.

\section{RELATED WORK}

\subsection{Precision Robotics Manipulation}
Previous works have considered robotic manipulations besides grasping such as stacking \cite{popov2017data}\cite{nair2018overcoming}, insertion \cite{thomas2018learning}\cite{vevcerik2017leveraging}\cite{vecerik2019practical}, screwing and kitting. These tasks require higher levels of precision with a slimmer margin for error, often necessitating extra algorithmic or sensory innovations. Some works focused on developing algorithms, in simulation \cite{nair2018overcoming} for precise robot manipulation tasks. Many of these methods leverage important state information such as accurate object poses that are available in simulation but difficult to obtain in the real world.
While this approach of using ground truth pose information of relevant entities is useful for developing algorithms for low-dimensional input space, generalizing such systems to the real world requires accurate pose estimation and a well calibrated system. Our approach learns directly from images and does not depend on such calibrated conditions.

In order to use algorithms for low-dimensional input space, some methods use fiducial markers to obtain pose information about objects in the scene \cite{thomas2018learning}. Other methods can be utilized to reason about object geometry \cite{yan2018learning}\cite{varley2017shape}, detect object pose \cite{chen2019grip}\cite{papazov2010efficient}\cite{litvak2019learning}\cite{tremblay2018deep}\cite{schmidt2014dart}\cite{andrychowicz2018learning}, key points \cite{manuelli2019kpam} and grasping points \cite{ten2017grasp} from pointcloud and RGB observations after which robot actions are planned and executed to accomplish tasks. The approaches require an estimate of both where a point in the world is relative to a camera, and where the camera is in relation to the robot.
Objects that are small, articulated, reflective, or transparent can complicate these methods.
In contrast, our method learns precision tasks in an end-to-end fashion without any intermediate pose estimation or camera calibration. Our approach learns hand-eye coordination across multi cameras, without requiring explicit pose information. With objects and the target goal visible from RGB view(s), the robot coordinates \textbf{\textit{relative}} displacements from its current pose rather than commanding the arm to absolute coordinates in a fixed reference frame.

Use of extra non-visual sensor modalities such as force-torque and tactile sensors is a common approach to enabling precision-based robotic tasks \cite{kalakrishnan2011learning}\cite{lange2012control}\cite{luo2019reinforcement}\cite{izatt2017tracking}\cite{watkins2019multi}.
These methods leverage contact-rich interactions with the environment to reactively achieve the desired manipulation.
Our vision-based approach is orthogonal and complementary to the use of non-visual inputs as demonstrated by hybrid methods that combine visual and non-visual sensing \cite{vecerik2019practical}\cite{lee2019making}\cite{jain2019learning}.

\begin{figure*}[t]
\vspace{2mm}
    \centering
    \includegraphics[width=\linewidth]{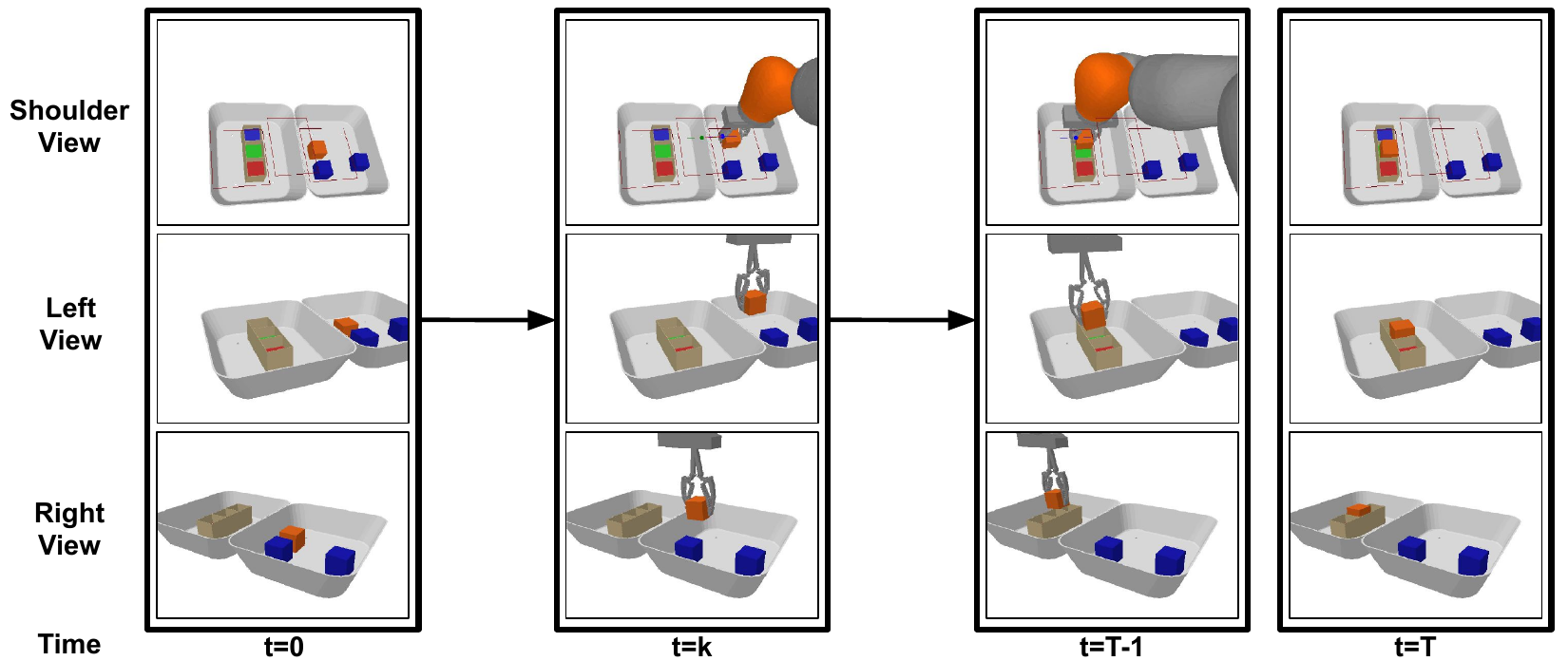}
    \caption{\small \textbf{Multi-view Insertion Task Learning.} A few key stages of the insertion task shown above include; start, pick, align, and drop. A single-view system that uses only the top image view struggles with stages that require 3D alignment. Our system combines information from multiple views and enhances performance on precision-based robotics. These 3 camera viewpoints are used for the Stacking I, Stacking II and Insertion tasks in the experimental section.}
    \label{insertion_task_large_figure}
\vspace{-4mm}
\end{figure*}

\begin{figure}[t]
\vspace{2mm}
    \centering
    \begin{subfigure}[b]{0.5\textwidth}
    \includegraphics[width=0.925\linewidth]{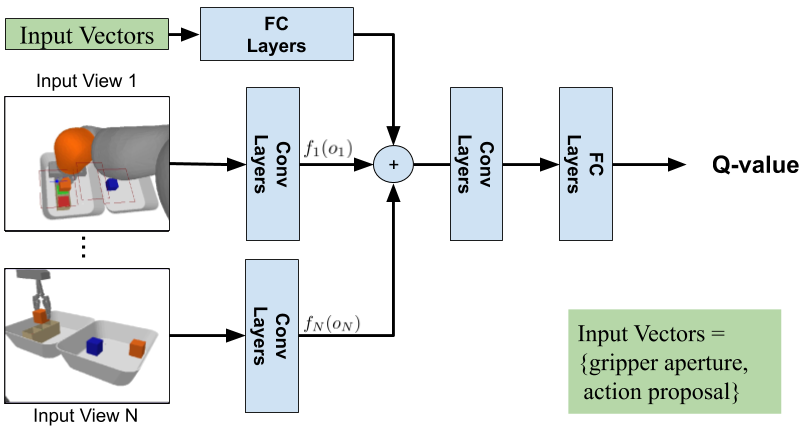}
    \caption{\small Multi-view Multi-Tower Q-Network Architecture}
    \label{Multi_Tower_Architecture}
        \end{subfigure}
    \begin{subfigure}[b]{0.5\textwidth}
        \includegraphics[width=0.925\linewidth]{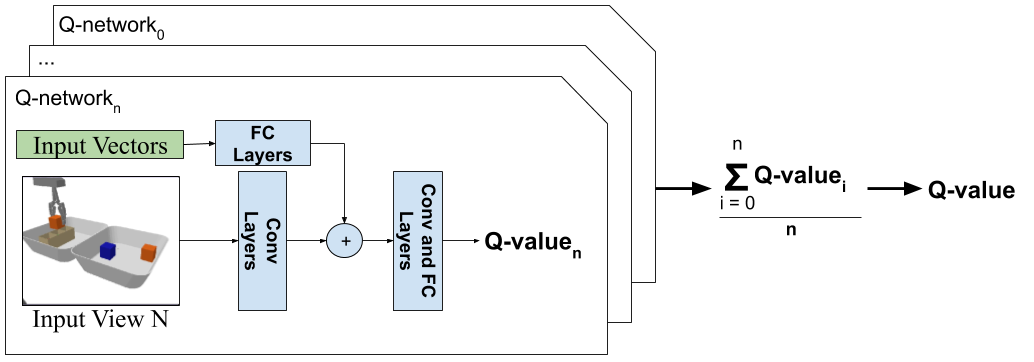}
    \caption{\small Multi-view Aggregate Q-Network Architecture}
    \label{Aggregate_Architecture}
    \end{subfigure}
    \caption{\small a) A multi-tower architecture for incorporating multiple views.  Each view has its own tower whose representations are then combined followed by additional network layers to produce a single Q-value.  b) An aggregate architecture has a separate Q-network for each individual view, the final Q-value is the mean of the per view Q-values. See Figure 13 of QT-Opt\cite{kalashnikov2018qt} for details of the single-view architecture (with Conv and FC block definitions) from which our multi-view architectures were adapted. We use their original single-view network with modified input vectors shown above as a baseline.
    }
\vspace{-5mm}
\end{figure}

\subsection{Vision-based Robot Manipulation}
Many learning-based methods can now perform robot manipulation such as grasping directly from high-dimensional input such as images.
For example, deep reinforcement learning (Deep-RL) algorithms can learn highly expressive neural networks by trial and error, that map image inputs to robot action.
One important consideration in learning-based systems like Deep-RL is the choice of environmental state observation; usually only a partial observation of the state is possible.
Color (RGB) image, depth image, 3D voxelized scene are all possible observation choices each with trade offs.
For example, voxel representations give rich 3D information to a particular resolution level but it are difficult to obtain directly from existing sensors, it is very high dimensional and can be inefficient and limiting when used in learning. Depth image gives 2.5D information which is more efficient but depth cameras still struggle to handle textureless, dark-colored, transparent or reflective materials depending on the technology used. RGB images give full color information useful for semantic understanding but lack depth information.
RGBD combines the advantage of the previous two, but the 3D understanding is partial (i.e. 2D) and is still susceptible to occlusion.


Generative Query Network (GQN) \cite{eslami2018neural} shows that a full scene can be represented using a vector encoding of multi-view images. Time Contrastive Networks (TCN) \cite{sermanet2018time} demonstrate that multiple views can be utilized to learn rich viewpoint invariant representations. Both GQN and TCN representations can be utilized to learn robotic manipulation. While these works are similar to ours in the use of multi-view capture of the scene, our approach does not use auxiliary loss functions. Rather, we allow the neural network to focus on extracting features relevant solely to the task at hand, rather than reconstructing the scene, or producing viewpoint invariant representations in a way that may hurt asymptotic performance on the task\cite{eslami2018neural}.

Some recent works \cite{gualtieri2017viewpoint}\cite{cheng2018reinforcement} use active sensing to determine successive sensor placements to improve state estimation for a task. In contrast, we use passive sensing via multiple cameras in a way that enables closed-loop reactive policy without adding intermediate camera-placement decision point into the sense-think-act loop.
A previous work \cite{mivseikis2016multi} used multiple static 3D cameras to capture and reconstruct the robot's environment in a non-learning based work. Our work differs in that we learn end-to-end directly from RGB images to robot action without any intermediate scene registration or reconstruction.

In summary, our approach addresses many limitations of these existing approaches to vision-based robot learning. Using an uncalibrated multi-camera system to capture RGB images from multiple viewpoints, we make the underlying state more observable and less susceptible to occlusion without suffering the computational and memory cost of an explicit 3D scene representation. Similar to \cite{kalashnikov2018qt}\cite{levine2018learning}, our method achieves closed-loop hand-eye coordination by learning the spatial relationship between the gripper and objects in the workspace, with reactive behaviors that ensure task success. While our experimental section is simulation only, there are  variety of works that demonstrate how our multi-view system could be deployed on real hardware \cite{kalashnikov2018qt}\cite{andrychowicz2018learning}\cite{james2019sim}\cite{chebotar2019closing}.

\section{PRELIMINARIES}
While our method would work with most existing imitation learning and reinforcement learning approaches, we adopt QT-Opt \cite{kalashnikov2018qt}-- a recent reinforcement learning algorithm that achieved state-of-the-art performance on a closed-loop vision-based robotic grasping task-- as the foundation for our approach. Here, we briefly summarize the RL markov decision process (MDP) and QT-Opt formulation, for additional details please see \cite{kalashnikov2018qt}.

\subsection{RL Formulation for Task Learning}
We use the standard RL MDP where the state of the task environment be given as $s_{t} \in \mathcal{S}$, the robot agent can make observation $o_{t} \in \mathcal{O}$ and can take action $a_{t} \in \mathcal{A}$ such that the transition dynamics is given as $s_{t+1}= \mathcal{T}(s_{t}, a_{t})$. For notational convenience, state and observation are used interchangeably in the MDP.
Given a reward function $\mathcal{R}(s_{t}, a_{t})$ that captures the desired behavior and a discount factor $\gamma\in [0,1)$, the goal of RL is to maximize the expected discounted cumulative reward across the episode.
Similar to QT-Opt, we use the sparse binary reward signal in our setup which indicates task success at the end of the episode and zero otherwise. In addition, we have a small constant negative reward at each time step to incentivise the agent to solve the task faster.
The action space includes 3D gripper displacement $\in \mathcal{R}^3$ and three binary commands (each $\in \{0, 1\}$) for gripper-open, gripper-close commands and a termination command to end the episode.
Different from QT-Opt and most existing RL works, our observation space consists solely of multiple image views from uncalibrated statically-placed cameras and the gripper aperture (an "Opened vs Closed" boolean).

\subsection{Q Target Optimization (QT-Opt)}

QT-Opt\cite{kalashnikov2018qt} is a continuous Q-Learning approach that learns a Q-value function which is then optimized, in a model predictive control fashion, to choose optimal actions that maximize the learned Q-function.
QT-Opt achieves this by learning a Q-function, represented by a neural network, that captures the expected discounted cumulative sum of reward starting from a given state and taking the action.

\begin{equation}\label{qlearning}
  \mathcal{Q}_{\theta}(s,a) = r(s,a) + \gamma  \max_{a'}\mathcal{Q}_{\theta}(s',a')
\end{equation}
The policy is recovered from a learned Q-function via: 
\begin{equation}\label{policy}
  \pi(s) = \underset{a}{\arg\max} \mathcal{Q}_{\theta}(s,a)
\end{equation}
The Q-function maximization is done with a cross-entropy method (CEM)-- a derivative-free optimization algorithm.

\begin{figure}[t]
\vspace{2mm}
    \centering
    \begin{subfigure}[b]{0.155\textwidth}
    \includegraphics[width=1\textwidth]{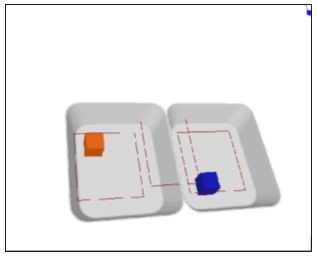}
    \caption{Stacking I: Start}
    \label{label:large_block_stacking_task_initial}
    \end{subfigure}
    \begin{subfigure}[b]{0.155\textwidth}
    \includegraphics[width=1\linewidth]{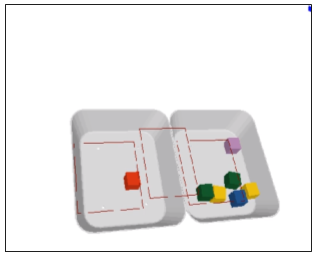}
    \caption{Stacking II: Start}
    \label{label:small_block_stacking_task_initial}
    \end{subfigure}
    \begin{subfigure}[b]{0.155\textwidth}
    \includegraphics[width=1\linewidth]{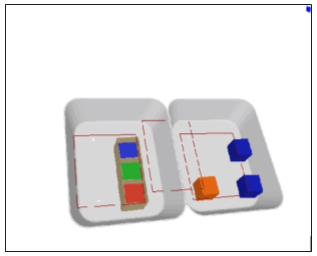}
        \caption{Insertion: Start}
        \label{label:insertion_task_initial}
    \end{subfigure}
    \begin{subfigure}[b]{0.155\textwidth}
    \includegraphics[width=1\linewidth]{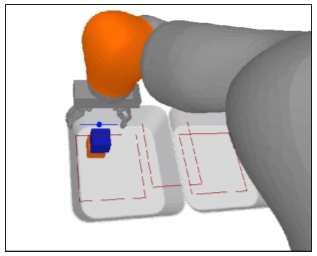}
        \caption{Stacking I: Goal}
        \label{label:large_block_stacking_task_final}
    \end{subfigure}
    \begin{subfigure}[b]{0.155\textwidth}
    \includegraphics[width=1\linewidth]{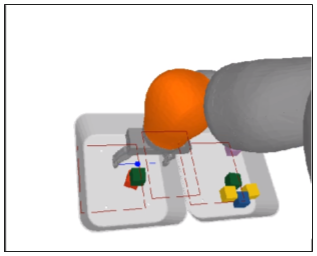}
        \caption{Stacking II: Goal}
        \label{label:small_block_stacking_task_final}
    \end{subfigure}
    \begin{subfigure}[b]{0.155\textwidth}
    \includegraphics[width=1\linewidth]{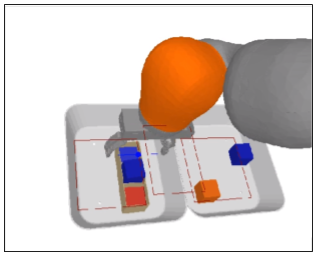}
        \caption{Insertion: Goal}
        \label{label:insertion_task_final}
    \end{subfigure}
    \caption{\small \textbf{Tasks with Varying Difficulty} The value of multi-view task learning depends on the level of 3D understanding and precision required for the task. The images above show sampled initial images and final images to illustrate the desired outcomes. \textbf{Left [a,d]}: The Stacking I task requires a block from the right side be placed on top of the block on the left side. The task has a large margin of error since the blocks are big enough that perfect alignment isn't required to succeed. \textbf{Middle [b,e]}: The blocks are smaller so there is a need for more precise placement, hence the performance benefit of having multiple views is potentially higher. \textbf{Right, [c,f]}: The insertion task requires the block placed into the middle placement location (green hole) of the fixture. This requires precise alignment which difficult from a single view, hence there is significant benefit to using multiple views.}
    \label{tasks}
\vspace{-5mm}
\end{figure}

\subsection{Robot Simulation Setup}
We use the Kuka IIWA arm with a parallel jaw gripper as the robot platform in our experiments, although our method is independent of the specific robot hardware. Using the Bullet Physics simulator \cite{coumans2016pybullet}, we create a simulation environment (Figure~\ref{insertion_task_large_figure}) where two bins are placed in front of the robot and based on the task at hand, objects are placed in the bins at random locations. Three cameras were mounted to overlook the bins where the task is being performed; with respect to the robot, one camera is over-the-shoulder, one is to the left and one is to the right.

\section{MULTI-VIEW TASK LEARNING}
Most vision-based learning algorithms take a single camera image input as an observation of the state of the environment. This may be reasonably sufficient for tasks requiring more coarse manipulation, but for tasks that require high-level of precision such as the insertion task in Figure \ref{insertion_task_large_figure}, a single view usually cannot capture enough of the state information to achieve superior performance.


For single-view task learning, we take a closer look at the Q-function expressed as a deep neural network and observe that it can be factorized given that the state input has visual and non-visual components i.e. $s = (s_{visual}, s_{non\_visual})$. The visual component is the image observation while the non-visual component is the gripper "Opened/Closed" status. As a result, the Q-function can be decomposed as:

\begin{equation} \label{eq1}
\begin{split}\mathcal{Q}_{\theta}(s,a) 
&= \mathcal{Q}_{\theta}(s_{visual}, s_{non\_visual},a) \\
&= \mathcal{Q}(f(s_{visual}), g(s_{non\_visual}), h(a)) \\
\end{split}
\end{equation}
where $f$, $g$ and $h$ are vector valued functions with $f$ being a sequence of 2D convolutional layers while $g$ and $h$ are a sequence of dense layers.

For a single view system, $f(s_{visual})$ is a function over the input image while in the multi-view setting, there can be different functions $f_1, f_2, ..., f_n$ that process image observations from viewpoints 1 through n. Below, we present different ways of expressing and composing the functions for both single and multiple views systems.

\subsection{Single View}

We evaluate against several single view architectures. Single View RGB from the shoulder (SV\_Shoulder).  We additionally explore the use of depth images from the shoulder view. We look at (SV\_RGBD) where the RGB and Depth are fed into separate CNN towers of the network. 

\subsection{Multi-Tower (MV\_Towers)}

Shown in Figure \ref{Multi_Tower_Architecture}, each image observation $o_i$ is passed through a separate vision processing module $f_i$ which is a sequence of convolution layers to produce visual embeddings $f_i(o_i)$ from each viewpoint. These embeddings are averaged across views and combined with action proposals before going through more convolution and dense layers to produce the Q-value $Q(s,a)$.
Note that even though the function for each viewpoint have the same form, each function is different with a separate set of weights i.e. $f_i \neq f_j,  \forall i \neq j$. The overall vision module is given as:
\[
f = \frac{1}{n}\sum_{i=1}^{n} f_i(o_i)
\]

\subsection{Siamese (MV\_Siamese)}
MV\_Siamese is a modification of MV\_Towers such that the convolutional weights across the multi-views are shared i.e. $f_i = f_j,  \forall i, j$. This reduces the total number of weights to be trained, shares the data from multiple views to train the same set of weights, imposes a constraint that visual data from all views should be processed similarly.

\subsection{Sensor Dropout (MV\_Dropout)}
MV\_Dropout is a modification of MV\_Towers. The data from one or more of the cameras is randomly masked out during training. This aims to improve robustness to the absence of camera views.
\[
f = \frac{1}{C} \sum_{i=1}^{n} \delta_i f_i(o_i)
\]
where $\delta_i \sim$ Bernoulli($p$) and $C =\sum_{i=1}^{n} \delta_i$ is the normalizing constant that ensures the scale of the output does not vary with the number of camera viewpoints selected.
Note that with this formulation it is possible to have no camera selected which is undesirable and a waste of training cycle even if happens less frequently.
An implementation detail is to list out all possible combination where one or more of the camera view-points are selected and put a uniform distribution to randomly select one of these possible options.
A similar sensor-dropout idea was shown to improve multi-sensory fusion for autonomous navigation task \cite{luo2019reinforcement}.

\subsection{Q-Aggregate Network (MV\_Q\_AGG)}
Shown in Figure \ref{Aggregate_Architecture} this multi-view approach creates a separate Q-function per input viewpoint and all Q outputs are aggregated into a single Q-value.
This is a consensus action approach where each view predicts the Q-value of each proposed action given the current state. The action with the highest Q-value (aggregated across views) is selected.
During training, the mean aggregate is preferred so that gradient flows through the entire network for all views; min/max operations would only allow gradient flow through a single branch of the network for each training datapoint. 
A drawback of the Q-Aggregate approach, regardless of the aggregate function, is that there are more parameters to train.

\begin{figure}[t] 
\vspace{2mm}
  \begin{minipage}[t]{1\linewidth}
    \centering
    \includegraphics[width=0.9\linewidth]{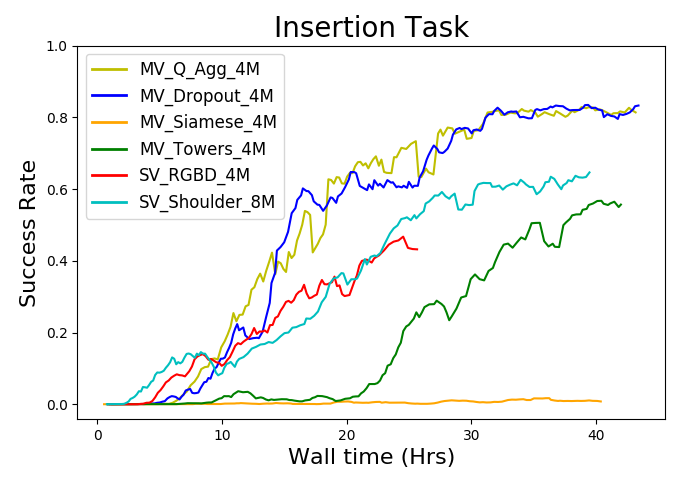}
    \caption{\small \textbf{Insertion Task Model Architecture}  Running average comparison of SV and MV architectures trained either 4 million (4M) or 8 million (8M) training iterations taking up to 40 hours. MV\_Dropout and MV\_Q\_Agg results achieve the best and comparable performance on this task. Importantly, sensor dropout during training leads to a huge difference in performance between MV\_Towers and MV\_Dropout.}
    \label{insertion_results_MV}
  \end{minipage}
\vspace{-5mm}
\end{figure}

\begin{figure}[t]
  \begin{minipage}[t]{1\linewidth}
    \centering
    \vspace{2mm}
    \includegraphics[width=0.875\linewidth]{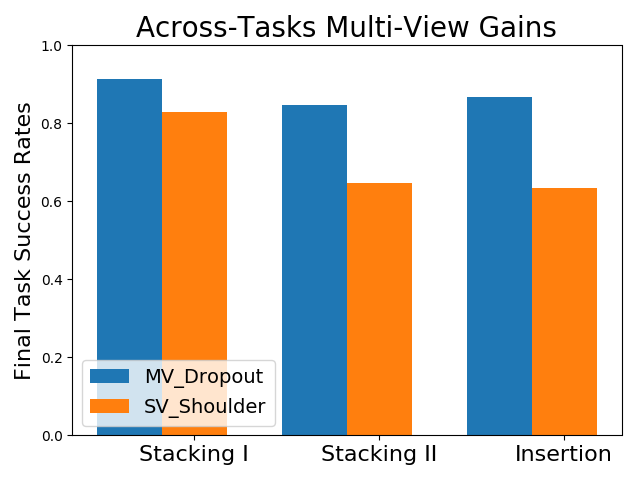}
    \caption{\small \textbf{Comparison of best performing SV and MV models across different tasks.} The relative gains from a multi-view approach are dependent on task with harder tasks gaining more.  Switching from single to multi-view results in the following absolute performance gains: Stacking I \textbf{8.97\%}; Stacking II \textbf{20.14\%}; and Insertion \textbf{23.43\%}}
    \label{across_task_performance_results}
  \end{minipage}
\vspace{-5mm}
\end{figure}

\begin{table*}[t]
\vspace{2mm}
\centering
\caption{\small \textbf{View Dropout Experiment} (\% Task Success on Insertion Task): This table overviews how different trained policies perform as the number of views available at runtime is reduced.  The Multi-View and Multi-View (Dropout) rows were all trained with observations from 3 views, but evaluated with 3, 2, and 1 view.  The single view baselines were trained with 1 view and evaluated with 1 view. Of note is the fact that the Multi-View (Dropout) significantly outperforms the Single View baselines even when only provided a single view at runtime.  It also outperforms the Multi-View model trained and evaluated with 3 views implying the dropout procedure has benefits even when all views are available. All numbers in the table come from the average of evaluating the final trained policy for 700 episodes.}
\label{tab:reliability_results}
\begin{tabular}{|c|c|c|c|c|c|c|c|c|}
\hline
\# of Views at Runtime: & 3 Views & \multicolumn{3}{|c|}{2 Views} & \multicolumn{3}{|c|}{1 View} \\\hline
& All Views & Shoulder + Left & Shoulder + Right &   Left + Right & Shoulder & Left & Right \\\hline
MV\_Towers\_4M [trained w/ 3 views]& 58.0 & 0.43 & 3.86 & 0 & 0 & 0 & 0 \\
MV\_Dropout\_4M [trained w/ 3 views] & \textbf{86.86} & 68.0 & 83.0 & 70.0 & 32.14 & 34.29 & 14.57 \\ \hline
SV\_Shoulder\_8M [trained w/ 1 view] & N/A & N/A & N/A & N/A & 63.43 & N/A & N/A  \\ \hline
\end{tabular}
\vspace{-3mm}
\end{table*}

\section{EXPERIMENTS}
For our experiments, three cameras were placed pointed at the robot's workspace.  With respect to the robot, one camera is roughly over the shoulder, one to the left and one to the right. For each episode, 0.01 std-dev uniform noise is added to the camera position, look and up vectors used to define the camera pose, for each camera, to simulate imperfect camera calibration and improve generalization. The bin locations are randomized with x, y, z position noise sampled uniformly from the ranges ($\pm 0.025$, $\pm 0.05$, $\pm 0.05$). To compare various multi-view and single-view approaches, three simulated robotic tasks were used as test-beds: 

\noindent\textbf{Stacking I} (Fig \ref{label:large_block_stacking_task_initial}, \ref{label:large_block_stacking_task_final}):  The right bin starts with a single block (5cm edge length) either blue or orange in a random position.  The left bin also starts with a single block either blue or orange.  An episode of the task is counted as a success if at the end of the episode there are two blocks in the left bin with one on top of the other.\\
\textbf{Stacking II} (Fig \ref{label:small_block_stacking_task_initial}, \ref{label:small_block_stacking_task_final}): The right bin starts with 6 small blocks (3.8cm edge length) in random positions, while the left bin starts with a single small cube in a random position.  An episode of the task is a success if at the end of the episode there are two blocks in the left bin with one on top of the other. In contrast to Stacking I, the blocks here are smaller which lowers the margin for error.\\
\textbf{Insertion} (Fig \ref{label:insertion_task_initial}, \ref{label:insertion_task_final}): The right bin starts with 3 blocks (5cm edge length) either blue or orange in random positions.  The left bin starts with a fixture at a random position, but fixed orientation.  An episode of the task is a success if at the end the fixture is in the left bin, and has a block firmly inserted into the middle fixture position. The fixture location has 9mm of clearance for the cube.

Poor exploration is a known issue in sparse-reward settings and previous works \cite{kalashnikov2018qt}\cite{zhu2018reinforcement} have shown the importance of providing some demonstration data to aid exploration and bootstrap the reinforcement learning. Similarly, we bootstrap off a simple scripted sub-optimal policy (yielding about $20\%$ for the Insertion task) to collect demonstration data which is included in the replay buffer to address the exploration issue.


Model performance is affected both by the number of training iterations and by the duration of time that training is run for. 180 data collection jobs are run for all training workflows producing $\sim$5000 episodes per hour for the insertion task with some variability across model performance and task. The 50,000 most recent episodes are kept in an experience replay buffer which takes $\sim$10 hours to initially reach capacity. Over time, the distribution of episodes in the buffer will shift as the policy learns the task and older episodes are evicted. We used 4 million gradient updates for the multi-view and RGBD architectures, but provided the SV\_Shoulder 8 million gradient steps in order to give it a comparable wall clock time and comparable amount of collected data. The models were trained following the QT-Opt setup with 1000 bellman update workers and 10 GPUs. 

\subsection{Single and Multi-View Insertion}
Figures \ref{label:insertion_task_initial} and \ref{label:insertion_task_final} shows the insertion task. As shown in Figure \ref{insertion_results_MV}, single view approaches to task learning struggle to learn this task where the margin for error in aligning the block to the fixture is very low.  This is true even when the single view approaches are provided depth information in the form of a depth image channel (SV\_RGBD). Conversely, we see the value of using multiple camera-view as input into the system. In addition, we see that using sensor dropout gives further improvement on the multi-view performance. This boost might be because dropping out camera views during training forces the network to squeeze out as much information from each camera. Note that at inference time, we use all three cameras.
The Q-Agg network also achieves comparable high performance although it contains more parameters.

\subsection{Varying Task Difficulty}
From Figure \ref{across_task_performance_results}, notice that the relative performance of the multi-view system compared to single-view varies by task. For tasks that can accommodate a high margin of precision error (such as Stacking I where the blocks are large enough to overlap with little precision), the performance gap between single versus multiple view systems is small. As tasks require more and more precise manipulation, the benefit of multiple views becomes more apparent. 
By switching from SV\_Shoulder to MV\_Dropout, task failure rate on  Stacking I dropped from 17.14\% to 8.71\%- a \textbf{49.18\%} reduction in failure rate i.e. (17.14\% - 8.71\%) / 17.14\%. Higher performance gains were seen on relatively harder tasks: Stacking II failure rate dropped from 35.43\% to 15.29\% a \textbf{56.84\%} reduction in failure rate and  Insertion dropped from 36.57\% to 13.14\% a \textbf{64.1\%} reduction in failure rate.





\subsection{Camera Dropout Robustness Test}
We experimentally demonstrate the robustness that result from the use of sensor dropout during training by randomly taking out one or two of the cameras during evaluation and measure the task performance. As show in Table \ref{tab:reliability_results}, while both multi-view approaches perform better that the single-view baseline, using sensor dropout during training makes a multi-view system robust to loss of camera after deployment; it leverages the inherent redundancy of the multi-view architecture to achieve a more reliable system.
Compared to MV\_Towers where loss of camera view after training is catastrophic, the performance of MV\_Dropout drops $\>4\%$ to $\>19\%$ when one view is taken out and $\>53\%$ to $\>73\%$ when two viewpoints are taken out depending on which views are removed.  The Shoulder and Left views are positioned close to each other so the Shoulder + Right performs much better than Shoulder + Left.  From the right view alone it can be difficult to see the fixture so it is the worst performing single view. The MV\_Dropout model is able to work half as well as the SV\_Shoulder model, when operating off only a Shoulder view, this model is not specialized to that specific image and also has had to learn to operate off other view combinations.

\section{CONCLUSION}
This paper introduces a multi-view approach to robotic learning for precision-based tasks.
Our approach directly learns the task without going through an intermediate step of reconstructing the scene. The effective use of multiple views enables a richer observation of the underlying state relevant to the task.
Experiments on precision-based stacking and insertion tasks show that our sensor-dropout approach to multi-view task learning achieves superior performance compared to the common single view approach. This improvement can be seen in the asymptotic performance as well as robustness to occlusion and loss of camera views. Our multi-view approach enables 3D tasks from RGB cameras without the need for explicit 3D representations and without camera-camera and camera-robot calibration.
In the future, similar multi-view benefits can be achieved with a single mobile camera by learning a camera placement policy in addition to the task policy.







\section*{ACKNOWLEDGMENT}
We thank Vikas Sindhwani for valuable feedback on the project and paper draft.

\bibliographystyle{IEEEtran}

\end{document}